\documentclass[lettersize,journal]{IEEEtran}
\usepackage{amsmath,amsfonts}
\usepackage{algorithmic}
\usepackage{algorithm}
\usepackage{array}
\usepackage[caption=false,font=normalsize,labelfont=sf,textfont=sf]{subfig}
\usepackage{textcomp}
\usepackage{stfloats}
\usepackage{url}
\usepackage{verbatim}
\usepackage{graphicx}
\usepackage{cite}
\usepackage{multirow}
\usepackage{stackengine}
\usepackage{makecell}
\usepackage{rotating}
\usepackage{xcolor}
\usepackage[justification=centering]{caption}

\newcommand\blfootnote[1]{%
	\begingroup
	\renewcommand\thefootnote{}\footnote{#1}%
	\addtocounter{footnote}{-1}%
	\endgroup
}

\begin{document}

\title{Skeletal Video Anomaly Detection using Deep Learning: Survey, Challenges and Future Directions}

\author{Pratik K. Mishra, Alex Mihailidis, Shehroz S. Khan
\thanks{Pratik K. Mishra, Alex Mihailidis, and Shehroz S. Khan are with the Institute of Biomedical Engineering, University of Toronto, Toronto, Canada, and also with the KITE – Toronto Rehabilitation Institute, University Health Network, Toronto, Canada (e-mail: pratik.mishra@mail.utoronto.ca; alex.mihailidis@utoronto.ca; shehroz.khan@uhn.ca).}
}

\markboth{Journal of \LaTeX\ Class Files,~Vol.~14, No.~8, August~2021}%
{Shell \MakeLowercase{\textit{et al.}}: A Sample Article Using IEEEtran.cls for IEEE Journals}


\maketitle

\begin{abstract}
The existing methods for video anomaly detection mostly utilize videos containing identifiable facial and appearance-based features. The use of videos with identifiable faces raises privacy concerns, especially when used in a hospital or community-based setting. Appearance-based features can also be sensitive to pixel-based noise, straining the anomaly detection methods to model the changes in the background and making it difficult to focus on the actions of humans in the foreground. Structural information in the form of skeletons describing the human motion in the videos is privacy-protecting and can overcome some of the problems posed by appearance-based features. In this paper, we present a survey of privacy-protecting deep learning anomaly detection methods using skeletons extracted from videos. We present a novel taxonomy of algorithms based on the various learning approaches. We conclude that skeleton-based approaches for anomaly detection can be a plausible privacy-protecting alternative for video anomaly detection. Lastly, we identify major open research questions and provide guidelines to address them.
\end{abstract}


\blfootnote{\copyright 2024 IEEE.  Personal use of this material is permitted.  Permission from IEEE must be obtained for all other uses, in any current or future media, including reprinting/republishing this material for advertising or promotional purposes, creating new collective works, for resale or redistribution to servers or lists, or reuse of any copyrighted component of this work in other works.}

\section{Introduction}
Anomalous events pertain to unusual or abnormal actions, behaviours or situations that can lead to health, safety and economical risks \cite{khan2014one}. Anomalous events, by definition, are largely unseen and not much is known about them in advance \cite{chandola2009anomaly}. Due to their rarity, diversity and infrequency, collecting labeled data for anomalous events can be very difficult or costly \cite{khan2014one, gautam2020minimum}. With the lack of predetermined classes and a few labelled data for anomalous events, it can be very hard to train supervised machine learning models \cite{khan2014one}. Therefore, a general approach in majority of anomaly detection algorithms is to train a model that can best represent the 'normal' events or actions, and any deviations from it can be flagged as an unseen anomaly \cite{mishra2021minimum}. 
Anomalous behaviours among humans can be attributed at an individual level (e.g., falls \cite{nogas2020deepfall}) or multiple people in a scene (e.g., pedestrian crossing \cite{li2013anomaly}, violence in a crowded mall \cite{boekhoudt2021hr}). In the context of video-based anomaly detection, the general approach is to train a model to learn the patterns of actions or behaviours of individual(s), background and other semantic information in the normal activities videos, and identify significant deviations in the test videos as anomalies. However, anomaly detection is a challenging task due to the lack of labels and often times the unclear definition of an anomaly \cite{chandola2009anomaly}. 

The majority of video-based anomaly detection approaches use RGB videos where the people in the scene are identifiable. While using RGB camera-based systems in public places (e.g., malls, airports) is generally acceptable, the situation can be very different in personal dwelling, community, residential or clinical settings \cite{senior2009protecting}.  In a home or residential setting (e.g., nursing homes), individuals or patients can be monitored in their personal space that may breach their privacy. The lack of measures to deal with the privacy of individuals can be a bottleneck in the adoption and deployment of the anomaly detection-based systems \cite{climent2021protection}. However, monitoring of people with physical, cognitive or aging issues is also important to improve their quality of life and care. Therefore, as a trade-off, privacy-protecting video modalities can fill that gap and be used in these settings to save lives and improve patient care. Wearable devices face compliance issues among certain populations, where people may forget or in some cases refuse to wear them \cite{ye2019challenges}. Some of the privacy-protecting camera modalities that has been used in the past for anomaly detection involving humans include depth cameras \cite{nogas2020deepfall, schneider2022unsupervised}, thermal cameras \cite{mehta2021motion}, and infrared cameras \cite{denkovski2022multi, kopuklu2021driver}. While these modalities can partially or fully obfuscate an individual’s identity, they require specialized hardware or cameras and can be expensive to be used by general population. Skeletons extracted from RGB camera streams using pose estimation algorithms provide a suitable solution of privacy protection over RGB and other types of cameras \cite{golda2022perception}. Skeleton tracking only focuses on body joints and ignores facial identity, full body scan or background information. 
The pixel-based features in RGB videos that mask important information about the scene are sensitive to noise resulting from illumination, viewing direction and background clutter, resulting in false positives when detecting anomalies \cite{luo2021normal}. Furthermore, due to redundant information present in these features (e.g., background), there is an increased burden on methods to model the change in those areas of the scene rather than focus on the actions of humans in the foreground. Extracting information specific to human actions can not only provide a privacy-protecting solution, but can also help to filter out the background-related noise in the videos and aid the model to focus on key information for detecting abnormal events related to human behaviour. 
The skeletons represent an efficient way to model the human body joint positions over time and are robust to the complex background, illumination changes, and dynamic camera scenes \cite{morais2019learning}. 
In addition to being privacy-protecting, skeleton features are compact, well-structured, semantically rich, and highly descriptive about human actions and motion \cite{morais2019learning}. Anomaly detection using skeleton tracking is an emerging area of research as awareness around privacy of individuals and their data grows. However, skeleton-based approaches may not be sufficient for situations that explicitly need facial information for analysis, including emotion recognition \cite{dhall2015video,taati2019algorithmic}, pain detection \cite{menchetti2019pain} or remote heart monitoring \cite{chen2018video}, to name a few.

\begin{sidewaystable*}
\scriptsize
\caption{Summary of reviewed papers.}
\label{tab:tab1}
\centering
\begin{tabular}{|l|l|l|l|l|l|l|l|l|l|l|}
\hline
\textbf{\makecell[l]{Learning\\approach}}         	& \textbf{Paper}                                & \textbf{Datasets used}             & \textbf{\makecell[l]{Experimental\\Setting}}		& \textbf{\makecell[l]{Number of\\people in\\scene}} & \textbf{\makecell[l]{Type of anomalies}}                               & \textbf{\makecell[l]{Pose\\estimation\\algorithm}} & \textbf{Model input}                                            & \textbf{Model type}                               		& \textbf{Anomaly score}                                                       & \textbf{\makecell[l]{Eval. metric\\AUC(ROC)\\(or other)}}          \\ \hline
\multirow{5}{*}{Reconstruction}     & Gatt et al. \cite{gatt2019detecting}           & UTD-MHAD         & Indoor 			& Single                             & \makecell[l]{Irregular body\\postures}                  				& \makecell[l]{Openpose,\\Posenet}        & \makecell[l]{Skeleton\\keypoints}			& \makecell[l]{1DConv-AE,\\LSTM-AE}                       	& \makecell[l]{Reconstruction\\error}                                                         & \makecell[l]{AUC(PR)=0.91,\\F score=0.98}           \\ \cline{2-11}
& Temuroglu et al. \cite{temuroglu2020occlusion} & Custom              & Outdoor         					& Multiple                           & \makecell[l]{Drunk walking}      				& Openpose                               & \makecell[l]{Skeleton\\keypoints}                                              					& AE                                                		& \makecell[l]{Reconstruction\\error}                                              & \makecell[l]{Average of\\recall and\\specificity=0.91}   \\ \cline{2-11}
& Suzuki et al. \cite{suzuki2021skeleton}        & Custom            & \makecell[c]{---}                       			& Single                             & \makecell[l]{Poor body\\movements in\\children}                           & Openpose                               & \makecell[l]{Motion time-\\series images} 	& CAE                                               		& \makecell[l]{Reconstruction\\error}                                                         & \makecell[l]{Accuracy=99.3,\\F score=0.99}         \\ \cline{2-11}
& Jiang et al. \cite{jiang2022deep}              & Custom          & Outdoor                                      			& Multiple                           & \makecell[l]{Abnormal pedestrian\\behaviours at\\grade crossings}        & Alphapose                              & \makecell[l]{Skeleton\\keypoints}                                              					& \makecell[l]{GRU Encoder-\\Decoder}                               		& \makecell[l]{Reconstruction\\error}            & 0.82            \\ \cline{2-11}
& Song et al. \cite{song2023analysis}         & Custom          & Outdoor         			& Multiple      & \makecell[l]{Abnormal pedestrian\\behaviours at\\grade crossings}        & Openpose                              & \makecell[l]{Skeleton\\keypoints}  					& GAN                               		& \makecell[l]{Discriminator\\score}            & 0.89            \\ \cline{2-11}
& Fan et al. \cite{fan2022video}                 & \makecell[l]{CUHK Avenue,\\UMN}         & \makecell[l]{Indoor and\\Outdoor}       		& Multiple                           & \makecell[l]{Anomalous human\\behaviours}                                & Alphapose                              & \makecell[l]{Video frame,\\Skeleton\\keypoints}                        			& \makecell[l]{GAN}                                         		& \makecell[l]{Reconstruction\\error of\\video frame}                                     & \makecell[l]{0.88\\0.99}                 \\ \hline
\multirow{5}{*}{Prediction} 					& Rodrigues et al. \cite{rodrigues2020multi}     & \makecell[l]{IITB-Corridor,\\ShanghaiTech,\\CUHK Avenue}     & Outdoor  		& Multiple                           & \makecell[l]{Abnormal human\\activities}                 & Openpose                               & \makecell[l]{Skeleton\\keypoints}                                              					& \makecell[l]{Multi-timescale\\1DConv\\encoder-decoder}        	& \makecell[l]{Prediction error\\from different\\timescales}                                   & \makecell[l]{0.67\\0.76\\0.83}   \\ \cline{2-11}
& Luo et al. \cite{luo2021normal}                & \makecell[l]{ShanghaiTech,\\CUHK Avenue}           & Outdoor        		& Multiple                           & \makecell[l]{Irregular body\\postures}              				& Alphapose                              & \makecell[l]{Skeleton\\joints graph}                                           					& \makecell[l]{Spatio-Temporal\\GCN}   		& Prediction error                       & \makecell[l]{0.74\\0.87}        \\ \cline{2-11}
& Zeng et al. \cite{Zeng2023}                    & \makecell[l]{UCSD Pedestrian,\\ShanghaiTech,\\CUHK Avenue,\\IITB-Corridor}    & Outdoor  	& Multiple                           & \makecell[l]{Anomalous human\\behaviours}    & HRNet                                  & \makecell[l]{Skeleton\\joints graph}         & \makecell[l]{Hierarchical\\Spatio-Temporal\\GCN}         & \makecell[l]{Weighted sum of\\prediction errors\\from different\\levels}                     & \makecell[l]{0.98\\0.82\\0.87\\0.7}  \\ \cline{2-11}
& Fan et al. \cite{fan2021anomaly}               & \makecell[l]{ShanghaiTech,\\CUHK Avenue}          & Outdoor              	& Multiple                           & \makecell[l]{Anomalous human\\actions}                                   & Alphapose                              & \makecell[l]{Skeleton\\keypoints}                                              					& \makecell[l]{GRU feed forward\\network}                   & Prediction error                   & \makecell[l]{0.83\\0.92}         \\ \cline{2-11}
& Pang et al. \cite{pang2022predicting}          & \makecell[l]{ShanghaiTech,\\CUHK Avenue}           & Outdoor           	& Multiple                           & \makecell[l]{Anomalous human\\actions}                                   & Alphapose                              & \makecell[l]{Skeleton\\keypoints}                                              					& Transformer                                       		& Prediction error         & \makecell[l]{0.77\\0.87}      \\ \cline{2-11}
& Huang et al. \cite{huang2022hierarchical}          & \makecell[l]{ShanghaiTech,\\CUHK Avenue,\\IITB-Corridor}           & Outdoor           	& Multiple                           & \makecell[l]{Anomalous human\\behaviours}                                   & HRNet                              & \makecell[l]{Skeleton\\joints graph}                                              					& \makecell[l]{Spatio-Temporal\\Graph Transformer}         		& \makecell[l]{Max of prediction\\errors of local \&\\global graphs}         & \makecell[l]{0.83\\0.89\\0.77}      \\ \hline
\multirow{5}{*}{\makecell[l]{Reconstruction+\\Prediction}}        	& Morais et al. \cite{morais2019learning}        & \makecell[l]{ShanghaiTech,\\CUHK Avenue}            & Outdoor              	& Multiple                           & \makecell[l]{Anomalous human\\actions}                                   & Alphapose                              & \makecell[l]{Skeleton\\keypoints}                                              					& \makecell[l]{GRU Encoder-\\Decoder}                               		& \makecell[l]{Weighted sum of\\reconstruction and\\prediction errors}                        & \makecell[l]{0.73\\0.86}    \\ \cline{2-11}
& Boekhoudt et al. \cite{boekhoudt2021hr}        & \makecell[l]{ShanghaiTech,\\HR Crime}           & \makecell[l]{Indoor and\\Outdoor}      	& Multiple                           & \makecell[l]{Human and Crime\\related anomalies} 			& Alphapose                              & \makecell[l]{Skeleton\\keypoints}                                              					& \makecell[l]{GRU Encoder-\\Decoder}                               		& \makecell[l]{Weighted sum of\\reconstruction and\\prediction errors}                        & \makecell[l]{0.73\\0.6}      \\ \cline{2-11}
& Li and Zhang \cite{li2022video}                & ShanghaiTech                     & Outdoor                          			& Multiple                           & \makecell[l]{Abnormal pedestrian\\behaviours}                            & Alphapose                              & \makecell[l]{Skeleton\\keypoints}                                              					& \makecell[l]{GRU Encoder-\\Decoder}         	& \makecell[l]{Weighted sum of\\reconstruction and\\prediction errors}                        & 0.75              \\ \cline{2-11}
& Li et al. \cite{li2021human}                   & \makecell[l]{ShanghaiTech,\\CUHK Avenue}           & Outdoor               	& Multiple                           & \makecell[l]{Human-related\\anomalous events}                            & Alphapose                              & \makecell[l]{Skeleton\\joints graph}                                          					& \makecell[l]{GCAE with\\embedded LSTM}                    & \makecell[l]{Sum of max\\reconstruction and\\prediction errors}                              & \makecell[l]{0.76, EER=30.7\\0.84, EER=20.7} \\ \cline{2-11}
& Wu et al. \cite{wu2022confidence}              & \makecell[l]{ShanghaiTech,\\CUHK Avenue}            & Outdoor              	& Multiple                           & \makecell[l]{Abnormal human\\actions}                                    				& Alphapose                              & \makecell[l]{Skeleton\\joints graph,\\Confidence\\scores}               			& GCN                    	& \makecell[l]{Confidence score\\weighted sum of\\reconstruction,\\prediction and\\SVDD errors} &        \makecell[l]{0.77\\0.85}   \\ \cline{2-11}
& Luo et al. \cite{luo2022memory}              & \makecell[l]{ShanghaiTech,\\IITB-Corridor}            & Outdoor              	& Multiple                           & \makecell[l]{Human-related\\video anomalies}                                    				& \makecell[c]{---}                   & \makecell[l]{Skeleton\\joints graph}               			& \makecell[l]{Memory Enhanced\\Spatial-Temporal\\GCAE}                    	& \makecell[l]{Sum of max\\reconstruction and\\prediction errors} &        \makecell[l]{0.78\\0.69}   \\ \cline{2-11}
& Li et al. \cite{li2023human}              & \makecell[l]{ShanghaiTech,\\CUHK Avenue}            & Outdoor     	& Multiple      & \makecell[l]{Human-related\\anomalous events}    				& \makecell[c]{Alphapose}                   & \makecell[l]{Skeleton\\keypoints}               			& \makecell[l]{Memory-augmented\\GAN}                    	& \makecell[l]{Sum of max\\reconstruction and\\prediction errors} &        \makecell[l]{0.75, EER=31.7\\0.84, EER=22.6}   \\ \hline
\end{tabular}
\end{sidewaystable*}

\begin{sidewaystable*}
\scriptsize
\caption{Summary of reviewed papers (continued).}
\label{tab:tab1_cotd}
\centering
\begin{tabular}{|l|l|l|l|l|l|l|l|l|l|l|}
\hline
\textbf{\makecell[l]{Learning\\approach}}         	& \textbf{Paper}                                & \textbf{Datasets used}             & \textbf{\makecell[l]{Experimental\\Setting}}		& \textbf{\makecell[l]{Number of\\people in\\scene}} & \textbf{\makecell[l]{Type of anomalies}}                               & \textbf{\makecell[l]{Pose\\estimation\\algorithm}} & \textbf{Model input}                                            & \textbf{Model type}                               		& \textbf{Anomaly score}                                                       & \textbf{\makecell[l]{Eval. metric\\AUC(ROC)\\(or other)}}          \\ \hline
\multirow{3}{*}{\makecell[l]{Reconstruction+\\Clustering}}         	& Markovitz et al. \cite{markovitz2020graph}     & \makecell[l]{ShanghaiTech,\\NTU-RGB+D,\\Kinetics-250}         & \makecell[l]{Indoor and\\Outdoor}		& Multiple                           & \makecell[l]{Anomalous human\\actions}          & \makecell[l]{Alphapose,\\Openpose}      & \makecell[l]{Skeleton\\joints graph}           	& \makecell[l]{GCAE,\\Deep clustering}                    	& \makecell[l]{Dirichlet process\\mixture model\\score}                                        & \makecell[l]{0.75\\0.85\\0.74}   \\ \cline{2-11}
& Cui et al. \cite{cui2021prototype}             & ShanghaiTech                        & Outdoor               			& Multiple                           & \makecell[l]{Human pose\\anomalies}               				& \makecell[c]{---}                  & \makecell[l]{Skeleton\\joints graph}                                          					& \makecell[l]{GCAE,\\Deep clustering}                    	& \makecell[l]{Dirichlet process\\mixture model\\score}                      & 0.77                    \\ \cline{2-11}
& Liu et al. \cite{liu2022self}                  & \makecell[l]{ShanghaiTech,\\CUHK Avenue}           & Outdoor       		& Multiple                           & \makecell[l]{Anomalous human\\behaviours}                                & Alphapose                              & \makecell[l]{Skeleton\\joints graph}                                          					& \makecell[l]{GCAE,\\Deep clustering}                    	& \makecell[l]{Dirichlet process\\mixture model\\score}                   & \makecell[l]{0.79\\0.88}    \\ \cline{2-11}
& Chen et al. \cite{chen2023multiscale}                  & \makecell[l]{ShanghaiTech,\\CUHK Avenue}           & Outdoor       		& Multiple                           & \makecell[l]{Anomalous human\\behaviours}                                & Alphapose                              & \makecell[l]{Skeleton\\joints graph}                                          					& \makecell[l]{Multiscale spatial\\temporal attention\\GCN}        	& \makecell[l]{Skeleton cluster\\anomaly score}                   & \makecell[l]{0.76, EER=31.1\\0.88, EER=19.2}    \\ \cline{2-11}
& Yan et al. \cite{yan2023memory}           & \makecell[l]{ShanghaiTech,\\UCF-Crime,\\NTU-RGB+D}           & Outdoor       		& Multiple         & \makecell[l]{Anomalous human\\actions}                 & Openpose                & \makecell[l]{Skeleton\\joints graph}                    					& \makecell[l]{GCAE,\\Deep clustering}        	& \makecell[l]{Skeleton cluster\\anomaly score}                   & \makecell[l]{0.77\\0.76\\0.77}    \\ \hline
\multirow{2}{*}{Clustering}	& Yang et al. \cite{yang2022two}                 & \makecell[l]{UCSD Pedestrian 2,\\ShanghaiTech}           & Outdoor            	& Multiple                           & \makecell[l]{Anomalous human\\behaviours and\\objects}               & Alphapose                              & \makecell[l]{Skeleton\\joints graph,\\Numerical\\features}              			& GCN                                              		& \makecell[l]{Skeleton cluster +\\Object anomaly\\score}                        & \makecell[l]{0.93\\0.82}    \\ \cline{2-11}
 & Javed et al. \cite{javed2023learning}       & \makecell[l]{ShanghaiTech,\\UCF-Crime,\\NTU-RGB+D}    & Outdoor    & Multiple   & \makecell[l]{Anomalous human\\actions}               & \makecell[c]{---}      & \makecell[l]{Video frame,\\Skeleton\\joints graph}    & \makecell[l]{GCN,\\Deep clustering}   & \makecell[l]{Dirichlet process\\mixture model\\score}     & \makecell[l]{0.81\\0.86\\0.88}    \\ \hline
\makecell[l]{Iterative self-\\training}           	& Nanjun et al. \cite{li2022self}                & \makecell[l]{ShanghaiTech,\\CUHK Avenue}           & Outdoor                	& Multiple                           & \makecell[l]{Human-related\\anomalous events}                            & Alphapose                              & \makecell[l]{Skeleton\\joints graph,\\Numerical\\features}              			& GCN                                              		& \makecell[l]{Self-trained fully\\connected layers\\output}                                  &  \makecell[l]{0.72, EER=34.1\\0.82, EER=23.9} \\ \hline
\makecell[l]{Multivariate\\gaussian\\distribution}	& Tani and Shibata \cite{tani2022frame}          & ShanghaiTech                    & Outdoor             			& Multiple                           & \makecell[l]{Anomalous human\\behaviours}                                & Openpose                               & \makecell[l]{Skeleton\\joints graph}                                          					& \makecell[l]{GCN, Multivariate\\gaussian distribution} 	& \makecell[l]{Mahalanobis\\distance}            & 0.77             \\ \hline
\makecell[l]{Prompt-guided zero-\\shot learning}	& Sato et al. \cite{sato2023prompt}          & \makecell[l]{RWF-2000\\Kinetics-250}                    & Outdoor             			& Multiple                           & \makecell[l]{Abnormal-human\\behavior events}                                & \makecell[l]{PPN\\HRNet}                               & \makecell[l]{Skeleton\\keypoints}          					& \makecell[l]{Residual multilayer\\perceptron} 	& \makecell[l]{Joint probability\\score}            & \makecell[l]{Accuracy=90.3\\0.79}       \\ \hline
\end{tabular}
\end{sidewaystable*}

In recent years, deep learning methods have been developed to use skeletons for different applications, such as  action recognition \cite{song2021human}, medical diagnosis \cite{suzuki2021skeleton}, and sports analytics \cite{badiola2021systematic}. The use of skeletons for anomaly detection in videos is an under-explored area, and concerted research is needed
\cite{suzuki2021skeleton}.
The human skeletons can help in developing privacy-preserving solutions for private dwellings, crowded/public areas, medical settings, rehabilitation centers and long-term care homes to detect anomalous events that impact health and safety of individuals. Use of this type of approach could improve the adoption of video-based monitoring systems in homes and residential settings. However, there is a paucity of literature on understanding the existing techniques that use skeleton-based anomaly detection approaches. We identify this gap in the literature and present one of the first surveys on the recent advancements in using skeletons for anomaly detection in videos. We identified the major themes in existing work and present a novel taxonomy that is based on how these methods learn to detect anomalous events.
We also discuss the applications where these approaches were used to understand their potential in bringing these algorithms in a personal dwelling, or long-term care scenario.

\section{Literature Survey}
We adopted a narrative literature review for this work. The following keywords (and their combinations) were used to search for relevant papers – skeleton, human pose, body pose, body joint, anomaly detection, and video. These keywords were searched on scholarly databases, including Google Scholar, IEEE Xplore, Elsevier and Springer. We mostly reviewed papers between year 2016 to 2023; therefore, the list may not be comprehensive. In this review, we only focus on the recent deep learning-based algorithms for skeletal video anomaly detection and do not include traditional machine learning-based approaches. There are works \cite{boekhoudt2022spatial, du2017rpan} on detecting anomalous behaviour using supervised approaches, however, it is outside the scope of this review as it focuses on unsupervised anomaly detection approaches. 
We did not adopt the systematic or scoping review search protocol for this work; therefore, our literature review may not be exhaustive. However, we tried our best to include the latest development in the field to be able to summarize their potential and identify challenges. 
In this section, we provide a survey of skeletal deep learning video anomaly detection methods. We present a novel taxonomy to study the skeletal video anomaly approaches based on learning approaches into four broad categories, i.e., reconstruction, prediction, their combinations and other specific approaches. Table \ref{tab:tab1} and \ref{tab:tab1_cotd} provides a summary of 29 relevant papers, based on the taxonomy, found in our literature search. Unless otherwise specified, the values in the last column of the table refer to AUC(ROC) values corresponding to each dataset in the reviewed paper. Six papers use reconstruction approach, six papers use prediction approach, seven papers use a combination of reconstruction and prediction approaches, five papers use a combination of reconstruction and clustering approaches, and five papers use other specific approaches.

\subsection{Reconstruction Approaches} \label{lit_recons}
In the reconstruction approaches, generally, an autoencoder (AE) or its variant model is trained on the skeleton information of only normal human activities. During training, the model learns to reconstruct the samples representing normal activities with low reconstruction error. Hence, when the model encounters an anomalous sample at test time, it is expected to give high reconstruction error.

Gatt et al. \cite{gatt2019detecting} used Long Short-Term Memory (LSTM) and 1-Dimensional Convolution (1DConv)-based AE models to detect abnormal human activities, including, but not limited to falls, using skeletons estimated from videos of a publicly available dataset.
Temuroglu et al. \cite{temuroglu2020occlusion} proposed a skeleton trajectory representation that handled occlusions and an AE framework for pedestrian abnormal behaviour detection. The pedestrian video dataset used in this work was collected by the authors, where the training dataset was composed of normal walking, and the test dataset was composed of normal and drunk walking. The pose skeletons were treated to handle occlusions using the proposed representation and combined into a sequence to train an AE. They compared the results of occlusion-aware skeleton keypoints input with keypoints without occlusion flags, keypoint image heatmaps and raw pedestrian image inputs. The authors used average of recall and specificity to evaluate the models due to the unbalanced dataset and found that occlusion-aware input achieved the highest results.
Suzuki et al. \cite{suzuki2021skeleton} trained a Convolutional AE (CAE) on good gross motor movements in children and detected poor limb motion as an anomaly. Motion time-series images \cite{suzuki2020enhancement} were obtained from skeletons estimated from the videos of kindergarten children participants. The motion time-series images were fed as input to a CAE, which was trained on only the normal data. The difference between the input and reconstructed pixels was used to localize the poor body movements in anomalous frames.
Jiang et al. \cite{jiang2022deep} presented a message passing Gated Recurrent Unit (GRU) encoder-decoder network to detect and localize the anomalous pedestrian behaviours in videos captured at the grade crossing. The field-collected dataset consisted of over 50 hours of video recordings at two selected grade crossings with different camera angles. The skeletons were estimated and decomposed into global and local components before being fed as input to the encoder-decoder network. The localization of the anomalous pedestrians within a frame was done by identifying the skeletons with reconstruction error higher than the empirical threshold. They manually removed wrongly detected false skeletons as they claim that the wrong detection issue was observed at only one grade crossing. However, an approach of manual removal of false skeletons is impractical in many real world applications where the data is very large, making the need of an automated false skeleton identification and removal step imperative.
In their following work \cite{song2023analysis}, the authors improved the performance of detecting abnormal pedestrian behaviors at grade crossings using a generative adversarial network (GAN)-based framework. Two LSTM-based branches within the generator were used to analyze both local and global motion patterns simultaneously, reconstructing the corresponding inputs in the temporal domain. The discriminator was a fully connected neural network and produced a score representing the likelihood of inputs being an anomaly.
Fan et al. \cite{fan2022video} proposed an anomaly detection framework which consisted of two pairs of generator and discriminator. The generators were trained to reconstruct the normal video frames and the corresponding skeletons, respectively. The discriminators were trained to distinguish the original and reconstructed video frames and the original and reconstructed skeletons, respectively. The video frames and corresponding extracted skeletons served as input to the framework during training; however, at test time, decision was made based on only reconstruction error of video frames.

\paragraph*{Challenges}
AEs or their variants are widely used in many video-based anomaly detection methods \cite{nogas2020deepfall}. The choice of the right architecture to model the skeletons is very important. Further, being trained on the normal data, they are expected to produce higher reconstruction error for the abnormal inputs than the normal inputs, which has been adopted as a criterion for identifying anomalies. However, this assumption does not always hold in practice, that is, the AEs can generalize well that it can also reconstruct anomalies well, leading to false negatives \cite{gong2019memorizing}.

\subsection{Prediction Approaches} \label{lit_pred}
In prediction approaches, a network is generally trained to learn the normal human behaviour by predicting the skeletons at the next time step(s) using the skeletons representing normal human actions at past time steps. During testing, the test samples with high prediction errors are flagged as anomalies as the network is trained to predict only the skeletons representing normal actions.

Rodrigues et al. \cite{rodrigues2020multi} suggested that abnormal human activities can take place at different timescales, and the methods that operate at a fixed timescale (frame-based or video-clip-based) are not enough to capture the wide range of anomalies occurring with different time duration. They proposed a multi-timescale 1DConv encoder-decoder network where the intermediate layers were responsible to generate future and past predictions corresponding to different timescales. The network was trained to make predictions on normal activity skeletons input. The prediction errors from all timescales were combined to get an anomaly score to detect abnormal activities.
Luo et al. \cite{luo2021normal} proposed a spatio-temporal Graph Convolutional Network (GCN)-based prediction method for skeleton-based video anomaly detection. The body joints were estimated and built into skeleton graphs, where the body joints formed the nodes of the graph. The spatial edges connected different joints of a skeleton, and temporal edges connected the same joints across time. A fully connected layer was used at the end of the network to predict future skeletons.
Zeng et al. \cite{Zeng2023} proposed a hierarchical spatio-temporal GCN, where high-level representations encoded the trajectories of people and the interactions among multiple identities while low-level skeleton graph representations encoded the local body posture of each person. The method was proposed to detect anomalous human behaviours in both sparse and dense scenes. The inputs were organized into spatio-temporal skeleton graphs whose nodes were human body joints from multiple frames and fed to the network. The network was trained on the input skeleton graph representations of normal activities. Optical flow fields and size of skeleton bounding boxes were used to determine sparse and dense scenes. For dense scenes with crowds, higher weights were assigned to high-level representations while for sparse scenes, the weights of low-level graph representations were increased. During testing, the prediction errors from different branches were weighted and combined to obtain the final anomaly score.
Fan et al. \cite{fan2021anomaly} proposed a  GRU feed-forward network that was trained to predict the next skeleton using past skeleton sequences and a loss function that incorporated the range and speed of the predicted skeletons.
Pang et al. \cite{pang2022predicting} proposed a skeleton transformer to predict future pose components in video frames and considered error between predicted pose components and corresponding expected values as anomaly score. They applied a multi-head self-attention module to capture long-range dependencies between arbitrary pairwise pose components and the temporal convolutional layer to concentrate on local temporal information.
Huang et al. \cite{huang2022hierarchical} proposed a spatio-temporal graph transformer to encode the hierarchical graph embeddings of human skeletons for jointly modeling the interactions between individuals and the correlations among body joints within a single individual. Input to the transformer was provided as global and local graphs. Each node in the global graph encoded the speed of an individual as well as the relative position and interaction relations between individuals. Each local graph encoded the pose of an individual.

\paragraph*{Challenges}
In these methods, it is difficult to choose how far in future (or past) the prediction should be made to achieve optimum results. This could potentially be determined empirically; however, in the absence of a validation set such solutions remain elusive. The future prediction-based methods can be sensitive to noise in the past data \cite{tang2020integrating}. Any small changes in the past can result in significant variation in prediction, and not all of these changes signify anomalous situations. 

\subsection{Combinations of learning approaches}
In this section, we discuss the existing methods that utilize a combination of different learning approaches, namely, reconstruction and prediction approaches, and reconstruction and clustering approaches.

\subsubsection{Combination of reconstruction and prediction approaches}
Some skeletal video anomaly detection methods utilize a multi-objective loss function consisting of both reconstruction and prediction errors to learn the characteristics of skeletons signifying normal behaviour and identify skeletons with large errors as anomalies.
Morais et al. \cite{morais2019learning} proposed a method to model the normal human movements in surveillance videos using human skeletons and their relative positions in the scene. The human skeletons were decomposed into two sub-components: global body movement and local body posture. The global movement tracked the dynamics of the whole body in the scene, while the local posture described the skeleton configuration. The two components were passed as input to different branches of a message passing GRU single-encoder-dual-decoder-based network. The branches processed their data separately and interacted via cross-branch message passing at each time step. Each branch had an encoder, a reconstruction-based decoder and a prediction-based decoder. The network was trained using normal data, and during testing, a frame-level anomaly score was generated by aggregating the anomaly scores of all the skeletons in a frame to identify anomalous frames. In order to avoid the inaccuracy caused by incorrect detection of skeletons in video frames, the authors left out video frames where the skeletons cannot be estimated by the pose estimation algorithm. Hence, the results in this work was not a good representation of a real-world scenario, which often consists of complex-scenes with occluding objects and overlapping movement of people.
Boekhoudt et al. \cite{boekhoudt2021hr} utilized the network proposed by Morais et al. \cite{morais2019learning} for detecting human crime-based anomalies in videos using a newly proposed crime-based video surveillance dataset.
Similar to the work by Morais et al. \cite{morais2019learning}, Li and Zhang \cite{li2022video} proposed a dual branch single-encoder-dual-decoder GRU network that was trained on normal behaviour skeletons estimated from pedestrian videos. The two decoders were responsible for reconstructing the input skeletons and predicting future skeletons, respectively. However, unlike the work by Morais et al. \cite{morais2019learning}, there was no provision of message passing between the branches.
Li et al. \cite{li2021human} proposed a single-encoder-dual-decoder architecture established on a spatio-temporal Graph CAE (GCAE) embedded with a LSTM network in hidden layers. The two decoders were used to reconstruct the input skeleton sequences and predict the unseen future sequences, respectively, from the latent vectors projected via the encoder. The sum of maximum reconstruction and prediction errors among all the skeletons within a frame was used as anomaly score for detecting anomalous frames. 
Wu et al. \cite{wu2022confidence} proposed a GCN-based encoder-decoder architecture that was trained using normal action skeleton graphs and keypoint confidence scores as input to detect anomalous human actions in surveillance videos. The skeleton graph input was decomposed into global and local components. The network consisted of three encoder-decoder pipelines: the global pipeline, the local pipeline and the confidence score pipeline. The global and local encoder-decoder-based pipelines learned to reconstruct and predict the global and local components, respectively. The confidence score pipeline learned to reconstruct the confidence scores. Further, a Support Vector Data Description (SVDD)-based loss was employed to learn the boundary of the normal action global and local pipeline encoder output in latent feature space. The network was trained using a multi-objective loss function, composed of a weighted sum of skeleton graph reconstruction and prediction losses, confidence score reconstruction loss and multi-center SVDD loss.
Luo et al. \cite{luo2022memory} proposed a single-encoder-dual-decoder memory enhanced spatial-temporal GCAE network, where spatial-temporal graph convolution was used to encode discriminative features of skeleton graphs in spatial and temporal domains. The memory module recorded patterns for normal behaviour skeletons. Further, the encoded representation was not fed directly into the reconstructing and predicting decoders but was used as a query to retrieve the most relevant memory items. The memory module was used to restrain the reconstruction and prediction capability of the network on anomalies.
Li et al. \cite{li2023human} proposed memory-augmented Wasserstein GAN with gradient penalty to predict future human skeleton trajectories from a given past and reconstruct the given past simultaneously. While the discriminator attempted to fit the Wasserstein distance between the distribution of real and generated samples, the generator tried to minimize the Wasserstein distance to draw the distribution of real and generated samples closer. A memory module was applied in the generator to mitigate the strong generalization ability.

\subsubsection{Combination of reconstruction and clustering approaches}
Some skeletal video anomaly detection methods utilize a two-stage approach to identify anomalous human actions using spatio-temporal skeleton graphs. In the first pre-training stage, a GCAE-based model is trained to minimize the reconstruction loss on input skeleton graphs. In the second fine-tuning stage, the latent features generated by the pre-trained GCAE encoder is fed to a clustering layer and a Dirichlet Process Mixture model is used to estimate the distribution of the soft assignment of feature vectors to clusters. Finally at the test time, the Dirichlet normality score is used to identify the anomalous samples.
Markovitz et al. \cite{markovitz2020graph} identified that anomalous actions can be broadly classified in two categories, fine and coarse-grained anomalies. Fine-grained anomaly detection refers to detecting abnormal variations of an action, e.g., abnormal type of walking. Coarse-grained anomaly detection refers to defining particular normal actions and regarding other actions as abnormal, such as determining dancing as normal and gymnastics as abnormal. They utilized a spatio-temporal GCAE to map the skeleton graphs representing normal actions to a latent space, which was soft assigned to clusters using a deep clustering layer. The soft-assignment representation abstracted the type of data (fine or coarse-grained) from the Dirichlet model. After pre-training of GCAE, the latent feature output of the encoder and clusters were fine-tuned by minimizing a multi-objective loss function consisting of both the reconstruction loss and clustering loss. They leveraged ShanghaiTech \cite{luo2017revisit} dataset to test the performance of their proposed method on fine-grained anomalies, and NTU-RGB+D\cite{shahroudy2016ntu} and Kinetics-250\cite{kay2017kinetics} datasets for coarse-grained anomaly detection performance evaluation. 
Cui et al. \cite{cui2021prototype} proposed a semi-supervised prototype generation-based method for video anomaly detection to reduce the computational cost associated with graph-embedded networks. Skeleton graphs for normal actions were estimated from the videos and fed as input to a shift spatio-temporal GCAE to generate features. It was not clear which pose estimation algorithm was used to estimate the skeletons from video frames. The generated features were fed to the proposed prototype generation module designed to map the features to prototypes and update them during the training phase. In the pre-training step, the GCAE and prototype generation module were optimized using a loss function composed of reconstruction loss and generation loss of prototypes. In the fine-tuning step, the entire network was fine-tuned using a multi-objective loss function, composed of reconstruction loss, prototype generation loss and cluster loss.
Later, Liu et al. \cite{liu2022self} used self-attention augmented graph convolutions for detecting abnormal human behaviours based on skeleton graphs. Skeleton graphs were fed as input to a spatio-temporal self-attention augmented GCAE and latent features were extracted from the encoder part of the trained GCAE. After pre-training of GCAE, the entire network was fine-tuned using a multi-objective loss function consisting of both the reconstruction loss and clustering loss.
Chen et al. \cite{chen2023multiscale} proposed a multiscale spatial temporal attention GCN, which included an encoder to extract features, a reconstruction decoder branch to optimize encoder, and a clustering layer branch to obtain anomaly scores. During training, the decoder is used to optimize the encoder by minimizing the reconstruction error. However, during testing, the decoder is discarded, and only the clustering layer is used to generate the anomaly score. It used three scales of human skeleton graphs, namely, joint, part and limb. Spatial attention graph convolution operation was carried out on each scale, and the output features of three scales were weighted and summed to constitute the multiscale skeleton features.
Yan et al. \cite{yan2023memory} proposed a deep memory storage clustering method based on GCAE to implement the real-time updating of pseudo-labels and network parameters. It consisted of a feature extraction, autoencoder, clustering, memory storage, self-supervision and scoring modules. The feature extraction module \cite{markovitz2020graph} and the autoencoder module were used to form the reconstructed pose sequence. The reconstructed sequence was then sent to the memory storage module for storage, and the soft cluster assignment was performed on each sample through the k-means clustering method \cite{khan2004cluster}. The autoencoder, clustering, and memory storage modules were used to update the pseudo-labels and network parameters iteratively.

\paragraph*{Challenges}
The combination-based methods can carry the limitations of the individual learning approaches, as described in Section \ref{lit_recons} and \ref{lit_pred}. Further, in the absence of a validation set, it is difficult to determine the optimum value of combination coefficients in a multi-objective loss function.

\subsection{Other Approaches}
This section discusses the methods that leveraged a pre-trained deep learning model to encode latent features from the input skeletons and used approaches such as, clustering and multivariate gaussian distribution, in conjunction for detecting human action-based anomalies in videos.

Yang et al. \cite{yang2022two} proposed a two-stream fusion method to detect anomalies pertaining to body movement and object positions. YOLOv3 \cite{redmon2018yolov3} was used to detect people and objects in the video frames. Subsequently, skeletons were estimated from the video frames and passed as input to a spatio-temporal GCN, followed by a clustering-based fully connected layer to generate anomaly scores for skeletons. The information pertaining to the bounding box coordinates and confidence score of the detected objects was used to generate object anomaly scores. Finally, the skeleton and object normality scores were combined to generate the final anomaly score for a frame. 
Nanjun et al. \cite{li2022self} used the skeleton features estimated from the videos for pedestrian anomaly detection using an iterative self-training strategy. The training set consisted of unlabelled normal and anomalous video sequences. The skeletons were decomposed into global and local components, which were fed as input to an unsupervised anomaly detector, iForest \cite{liu2012isolation}, to yield the pseudo anomalous and normal skeleton sets. The pseudo sets were used to train an anomaly scoring module, consisting of a spatial GCN and fully connected layers with a single output unit. As part of the self-training strategy, new anomaly scores were generated using previously trained anomaly scoring module to update the membership of skeleton samples in the skeleton sets. The scoring module was then retrained using updated skeleton sets, until the best scoring model was obtained. However, the paper doesn't discuss the criteria to decide the best scoring model.
Tani and Shibata \cite{tani2022frame} proposed a framework for training a frame-wise Adaptive GCN (AGCN) for action recognition using single frame skeletons and used the features extracted from the AGCN to train an anomaly detection model. As part of the proposed framework, a pretrained action recognition model \cite{shi2019two} was used to identify the frames with large temporal attention in the Kinetics-skeleton dataset \cite{yan2018spatial} as the action frames to train the AGCN. Further, the trained AGCN was used to extract features from the normal behaviour skeletons identified in the ShanghaiTech Campus dataset \cite{morais2019learning} to model a multivariate gaussian distribution. During testing, the Mahalanobis distance was used to calculate the anomaly score under the multivariate gaussian distribution.
Sato et al. \cite{sato2023prompt} proposed a user prompt-guided zero-shot learning framework for the detection of abnormal human behaviour events. A multilayer perceptron feature extractor was pretrained on large-scale action recognition datasets \cite{carreira2017quo,liu2019ntu} using contrastive learning between the skeleton features and the text embeddings extracted from action class names. The distribution of skeleton features of the normal actions was modeled during training while freezing the weights of feature extractor. During inference, the anomaly score was computed using distribution and the text prompts of an unseen action.
Javed et al. \cite{javed2023learning} proposed a unified framework for learning suitable frames of interest to cut down on redundant data and a two-stream feature block with a hyper-gated fusion model to take advantage of skeleton graph and video frame features. Soft assignments were later processed through a clustering layer, where probabilities were assigned to the instances and a normality score was calculated using the Dirichlet Process Mixture model \cite{blei2006variational}.

\paragraph*{Challenges}
The performance of these methods rely on the pre-training strategy of the deep learning models used to learn the latent features and the choice of training parameters for the subsequent machine learning models.

\begin{table*}
\caption{Characteristics of skeletal video anomaly detection datasets.}
\label{tab:datasets}
\centering
\begin{tabular}{|l|l|l|l|l|l|l|l|}
\hline
\textbf{Dataset} & \textbf{Total frames} & \textbf{\makecell[l]{Training\\frames}} & \textbf{Test frames} & \textbf{\makecell[l]{Anomalous\\events}} & \textbf{\makecell[l]{Camera\\views}} & \textbf{\makecell[l]{Available\\annotations}} & \textbf{Anomalies}                                                                                                          \\ \hline
\makecell[l]{CUHK\\Avenue\cite{lu2013abnormal}}      & 30652                 & 15328                    & 15324                & 47                        & 1               & \makecell[l]{Temporal,\\Pixel-wise,\\Track-ID} & \makecell[l]{Throwing object, child skipping, wrong\\direction, bag on grass}                                                              \\ \hline
IITB-Corridor\cite{rodrigues2020multi}    & 483566                & 301999                   & 181567               & \makecell[l]{108278\\frames}             & 1               & Temporal                       & \makecell[l]{Protest, unattended baggage, biker, fighting,\\chasing, loitering, suspicious object, hiding,\\playing with ball}               \\ \hline
ShanghaiTech\cite{luo2017revisit}     & 317398                & 274515                   & 42883                & 130                       & 13              & \makecell[l]{Temporal,\\Pixel-wise}           & \makecell[l]{Throwing object, jumping, pushing, bikers,\\loitering, climbing}                                                              \\ \hline
UCF-Crime\cite{sultani2018real}        & 1900 videos           & 1610 videos              & 290 videos           & 950 videos                & ---             & Temporal                       & \makecell[l]{Abuse, arrest, arson, assault, accident,\\burglary, explosion, fighting, robbery,\\shooting, stealing, shoplifting, vandalism} \\ \hline
\makecell[l]{UCSD\\Pedestrian\cite{li2013anomaly}}  & 18560                 & 9350                     & 9210                 & 77                        & 2               & \makecell[l]{Temporal,\\Pixel-wise,\\Track-ID} & \makecell[l]{Biker, skater, cart, wheelchair, walk across\\walkways}                                                                       \\ \hline
UMN\cite{umndataset}     & 3855                  & ---                      & ---                  & 11                        & ---             & Temporal                       & \makecell[l]{Abandoned or thrown objects, unusual\\crowd activity, camera tampering}                  \\ \hline
\end{tabular}
\end{table*}

\begin{figure*}[t]
    \centering
    \stackunder[2pt]{\includegraphics[width=5cm, height=3cm]{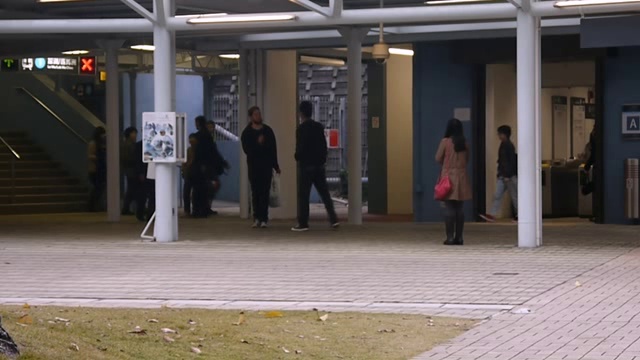}}{CUHK Avenue (Normal)}
    \stackunder[2pt]{\includegraphics[width=5cm, height=3cm]{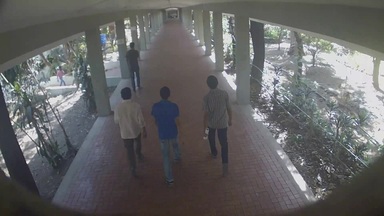}}{IITB-Corridor (Normal)}
    \stackunder[2pt]{\includegraphics[width=5cm, height=3cm]{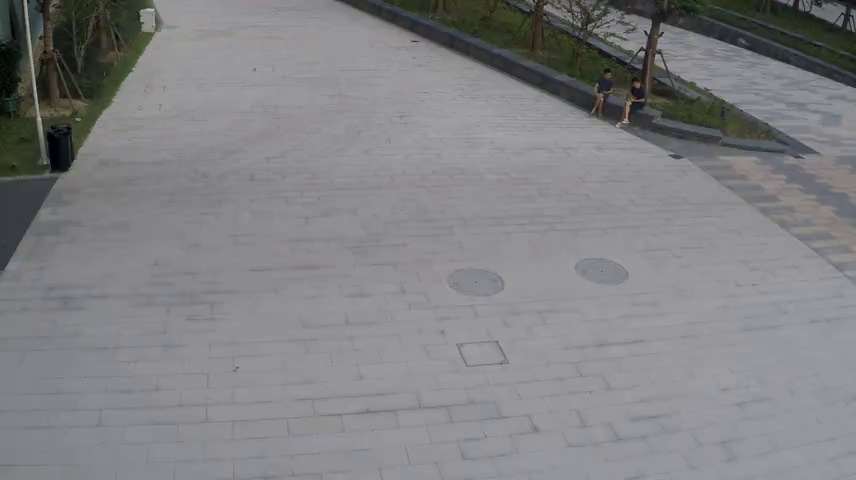}}{ShanghaiTech (Normal)}
    
    \stackunder[2pt]{\includegraphics[width=5cm, height=3cm]{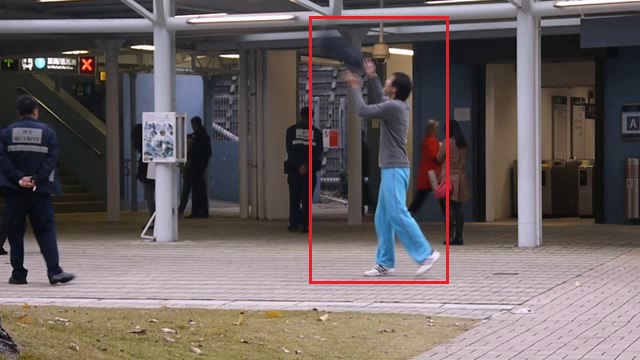}}{CUHK Avenue (Throwing object)}
    \stackunder[2pt]{\includegraphics[width=5cm, height=3cm]{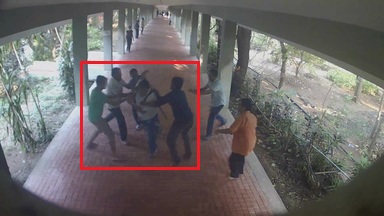}}{IITB-Corridor (Fighting)}
    \stackunder[2pt]{\includegraphics[width=5cm, height=3cm]{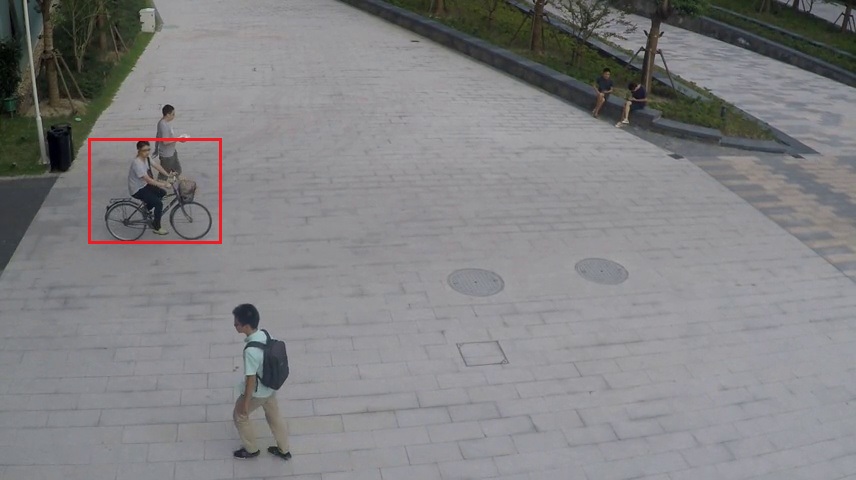}}{ShanghaiTech (Biker)}
    
    \stackunder[2pt]{\includegraphics[width=5cm, height=3cm]{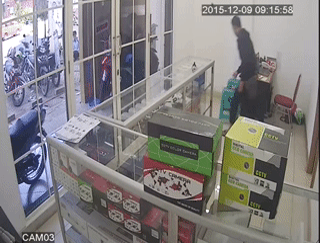}}{UCF-Crime (Normal)}
    \stackunder[2pt]{\includegraphics[width=5cm, height=3cm]{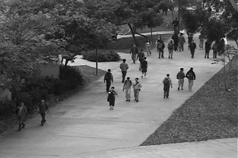}}{UCSD peds1 (Normal)}
    \stackunder[2pt]{\includegraphics[width=5cm, height=3cm]{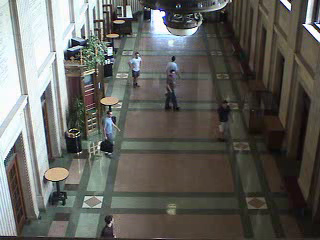}}{UMN (Normal)}
    
    \stackunder[2pt]{\includegraphics[width=5cm, height=3cm]{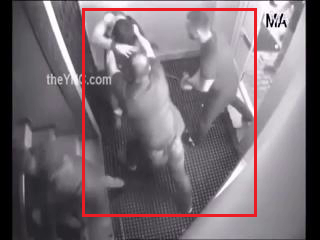}}{UCF-Crime (Assault)}
    \stackunder[2pt]{\includegraphics[width=5cm, height=3cm]{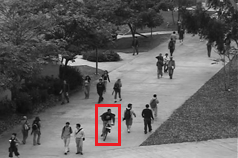}}{UCSD peds1 (Biker)}
    \stackunder[2pt]{\includegraphics[width=5cm, height=3cm]{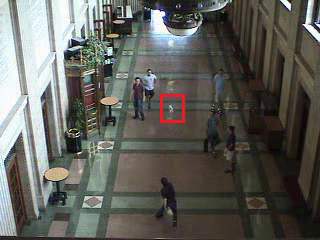}}{UMN (Throwing object)}
\caption{One normal and one anomalous frame from each of the skeletal video anomaly detection datasets.}
\label{fig:datasets}
\end{figure*}

\section{Discussion} \label{conc}
This section leverages Table \ref{tab:tab1} and \ref{tab:tab1_cotd} and synthesizes the information and trends that can be inferred from the existing work on skeletal video anomaly detection.

\subsection{Datasets}
ShanghaiTech \cite{luo2017revisit} and CUHK Avenue \cite{lu2013abnormal} were the most frequently used video datasets to evaluate the performance of the skeletal video anomaly detection methods. The ShanghaiTech dataset has videos of people walking along a sidewalk of the ShanghaiTech university and anomalous frames contain bikers, skateboarders and people fighting. It has 330 training videos and 107 test videos. However, not all the anomalous activities are related to humans. A subset of the ShanghaiTech dataset that contained anomalous activities only related to humans was termed as HR ShanghaiTech and was used in many papers. The CUHK Avenue dataset consists of short video clips looking at the side of a building with pedestrians walking by it. Concrete columns that are part of the building cause some occlusion. The dataset contains 16 training videos and 21 testing videos. The anomalous events comprise of actions such as “throwing papers”, “throwing bag”, “child skipping”, “wrong direction” and “bag on grass”. Similarly, a subset of the CUHK Avenue dataset containing anomalous activities only related to humans, called HR Avenue, has been used to evaluate the methods. Other video datasets that have been used include UTD-MHAD \cite{chen2015utd}, UMN \cite{umndataset}, UCSD Pedestrian\cite{li2013anomaly}, IITB-Corridor \cite{rodrigues2020multi}, UCF-Crime\cite{sultani2018real}, HR Crime\cite{boekhoudt2021hr}, NTU-RGB+D\cite{shahroudy2016ntu}, RWF-2000\cite{9412502} and Kinetics-250\cite{kay2017kinetics}.
Table \ref{tab:datasets} presents a summary of the characteristics of these datasets and Figure \ref{fig:datasets} presents one normal and one anomalous frame from these datasets. Among the datasets used in the reviewed papers, some of the datasets were originally not meant for but instead adopted for the task of video anomaly detection. Hence, we only provide details for the datasets that were originally meant for video anomaly detection in Table \ref{tab:datasets} and Figure \ref{fig:datasets}.
From the type of anomalies present in these datasets, it can be inferred that the existing skeletal video anomaly detection methods have been evaluated mostly on individual human action-based anomalies. Hence, it is not clear how well can they detect anomalies that involve interactions among multiple individuals or interactions among people and objects.

\subsection{Number of people in the scene}
Most of the papers (27 out of 29), detected anomalous human actions for multiple people in the video scene. Other two papers detected irregular body postures and poor body movements in children, respectively, for single person in the video scene. The usual approach was to estimate the skeletons for the people in the scene using a pose estimation algorithm, and calculate anomaly scores for each of the skeletons. The maximum anomaly score among all the skeletons within a frame was used to identify the anomalous frames. A single video frame could contain multiple people, among which not all of them were performing anomalous actions. Hence, taking the maximum anomaly score of all the skeletons helped to nullify the effect of people with normal actions on the final decision for the frame. Further, calculating anomaly scores for individual skeletons helped to localize the source of anomaly within a frame.

\subsection{Fields of application}
The definition of anomalous human behaviours can differ across various applications. While most of the existing papers focused on detecting anomalous human behaviours in general, five papers focused on detecting anomalous behaviours for specific applications, including drunk walking \cite{temuroglu2020occlusion}, poor body movements in children \cite{suzuki2021skeleton}, abnormal pedestrian behaviours at grade crossings \cite{jiang2022deep, song2023analysis} and crime-based anomalies \cite{boekhoudt2021hr}. Moreover, the nature of anomalous behaviours can vary depending upon various factors, such as span of time, crowded scenes, and specific action-based anomalies. Some papers identified and addressed the need to detect specific types of anomalies, namely, multi-timescale anomalies occurring over different time duration \cite{rodrigues2020multi}, anomalies in both sparse and crowded scenes \cite{Zeng2023}, fine and coarse-grained anomalies \cite{markovitz2020graph} and body movement and object position anomalies \cite{yang2022two}.

\subsection{Choice of pose estimation algorithm}
Alphapose \cite{fang2017rmpe} and Openpose \cite{cao2017realtime} were the most common choice of pose estimation algorithm for extraction of skeletons for the people in the scene. Other pose estimation methods that have been used were Posenet\cite{papandreou2017towards}, PPN\cite{sekii2018pose} and HRNet\cite{sun2019deep}. However, in general, the papers did not provide any rationale behind their choice of the pose estimation algorithm.

\subsection{Model type}
The type of models used in the papers can broadly be divided into two types, sequence-based and graph-based models. The sequence-based models that have been used include 1DConv-AE, LSTM-AE, GRU, and Transformer. These models treated skeleton keypoints for individual people across multiple frames as time series input. The graph-based models that have been used involve GCAE and GCN. The graph-based models received spatio-temporal skeleton graphs for individual people as input. The spatio-temporal graphs were constructed by considering body joints as the nodes of the graph. The spatial edges connected different joints of a skeleton, and temporal edges connected the same joints across time.

\subsection{Evaluation metrics}
The choice of a suitable threshold for anomaly detection can vary across different applications as most applications come with different costs for false alarms and missed anomalies \cite{ruff2021unifying, khan2016dtfall}. As such, having a metric capable of evaluating the performance of anomaly detection methods across diverse application scenarios, or equivalently, across a wide array of decision thresholds is highly desirable. The Area Under Curve (AUC) of Receiver Operating Characteristic (ROC) curve computes the fraction of detected anomalies, averaged over the full range of decision thresholds. It is the standard evaluation measure used in anomaly detection \cite{ruff2021unifying} and also the most common metric used to evaluate the performance among the existing skeletal video anomaly detection methods. The highest AUC(ROC) values reported for the commonly used ShanghaiTech \cite{luo2017revisit} and CUHK Avenue \cite{lu2013abnormal} datasets across different methods in Table \ref{tab:tab1} and \ref{tab:tab1_cotd} were 0.83 and 0.92, respectively. A direct comparison may not be possible due to the difference in the experimental setup and train-test splits across the reviewed methods; however, it gives some confidence on the viability of these approaches for skeletal video anomaly detection.
Other performance evaluation metrics include F score, accuracy, Equal Error Rate (EER) and AUC of Precision-Recall (PR) Curve. EER signifies the percentage of misclassified frames when the false positive rate equals to the miss rate on the ROC curve.
While AUC(ROC) can provide a good estimate of the classifier's performance over different thresholds, it can be misleading in case the data is imbalanced \cite{saito2015precision}. In anomaly detection scenario, it is common to have imbalance in the test data, as the anomalous behaviours occur infrequently, particularly in many medical applications \cite{galvao2021framework, 9684388}. The AUC(PR) value provides a good estimate of the classifier's performance on imbalanced datasets \cite{saito2015precision}; however, only one of the papers used AUC(PR) as an evaluation metric.

\begin{figure*}[t]
    \centering
    \stackunder[2pt]{\includegraphics[width=4.5cm, height=3cm]{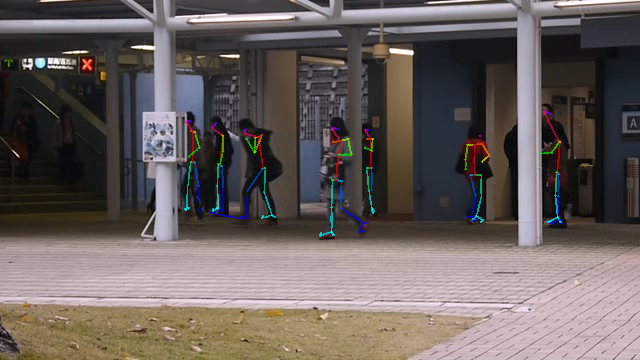}}{}
    \stackunder[2pt]{\includegraphics[width=4.5cm, height=3cm]{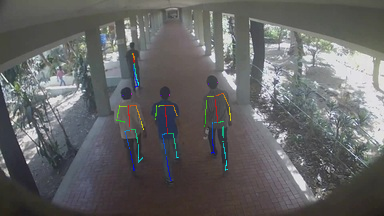}}{}
    \stackunder[2pt]{\includegraphics[width=4.5cm, height=3cm]{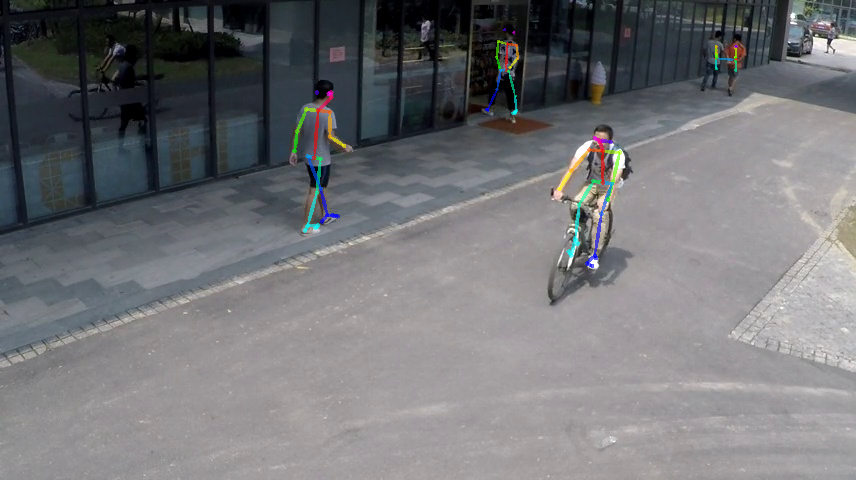}}{}
    
    \stackunder[2pt]{\includegraphics[width=4.5cm, height=3cm]{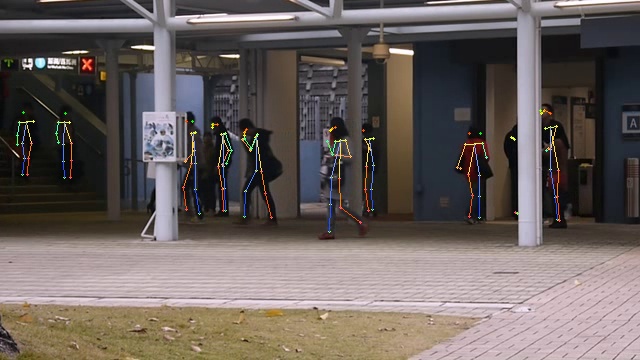}}{}
    \stackunder[2pt]{\includegraphics[width=4.5cm, height=3cm]{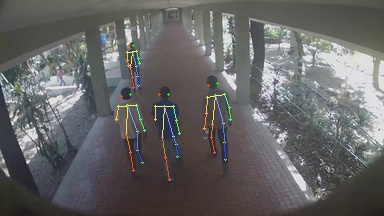}}{}
    \stackunder[2pt]{\includegraphics[width=4.5cm, height=3cm]{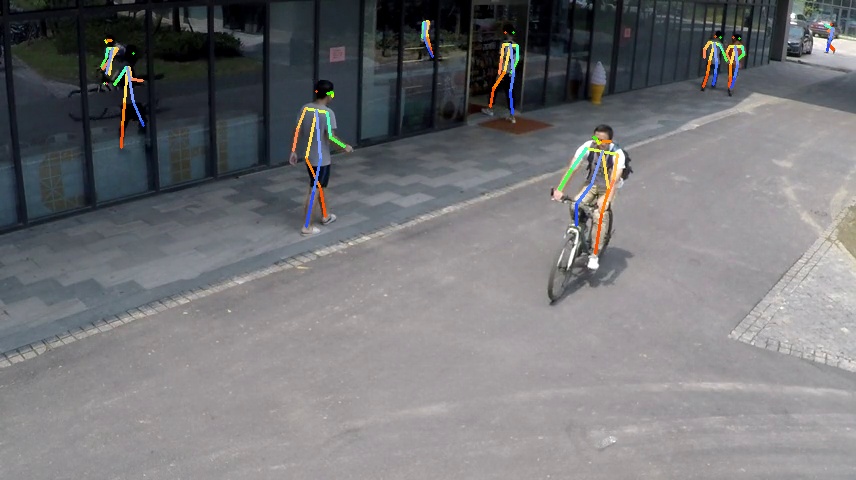}}{}
\caption{Openpose (top row) and alphapose (bottom row) output on different datasets.}
\label{fig:poses}
\end{figure*}

\section{Challenges}
\subsection{Pose estimation algorithms}
In general, the efficiency of the skeletal video anomaly detection algorithms depends upon the accuracy of the skeletons estimated by the pose-estimation algorithm. If the pose estimation algorithm misses certain joints or produces artifacts in the scene, then it can increase the number of false alarms. There are various challenges associated with estimating skeletons from video frames \cite{chen2020monocular}: (i) complex body configuration causing self-occlusions and complex poses, (ii) diverse appearance, including clothing, and (iii) complex environment with occlusion from other people in the scene, various viewing angles, distance from camera and truncation of parts in the camera view. This can lead to a poor approximation of skeletons and can negatively impact the performance of the anomaly detection algorithms. Further, there is an associated high cost of powerful hardware required for extracting skeletons using deep learning methods. Methods have been proposed to address some of these challenges \cite{cheng2019occlusion, gong2010learning}; however, extracting skeletons in complex environments remains a difficult problem. 
The two most commonly used pose estimation algorithms in the reviewed papers are Openpose \cite{cao2017realtime} and Alphapose \cite{fang2017rmpe}. Multi-person pose estimation can be categorized into top-down and bottom-up methods \cite{chen2020monocular}. Top-down methods \cite{fang2017rmpe, iqbal2016multi} usually employ human detectors to obtain bounding boxes for humans in the input frame and then utilize existing single-person pose estimators to predict body joints. This method highly depends upon the precision of human detection algorithms, and the run-time is proportional to the number of persons in the input frame. Bottom-up methods \cite{cao2017realtime} directly approximate all the body joints of all the humans in the input frame and assemble them into individual skeletons. However, the grouping of joints in a complex scene is a challenging task. Openpose is a bottom-up method and Alphapose is a top-down method. 
Figure \ref{fig:poses} presents the skeleton output of openpose and alphapose on different dataset frames.
Some of the existing methods manually remove inaccurate and false skeletons \cite{morais2019learning, jiang2022deep} to train the model, which is impractical in many real-world applications where the amount of available data is very large. There is a need for an automated false skeleton identification and removal step when estimating skeletons from videos. 

\subsection{Types of anomalies}
The anomalous human behaviours of interest and their difficulty of detection can vary depending upon the definition of anomaly, application, time span of the anomalous actions, and presence of single/multiple people in the scenes. For example, in the case of driver anomaly detection application, the anomalous behaviours can include talking on the phone, dozing off or drinking \cite{kopuklu2021driver}. The anomalous actions can span over different time lengths, ranging from few seconds to hours or days, e.g., jumping and falls \cite{khan2017detecting} are short-term anomalies, while loitering and social isolation \cite{boamah2021social} are long-term events. More focus is needed on developing methods that can identify both short and long-term anomalies. 
Sparse scene anomalies can be described as anomalies in scenes with less number of humans, while dense scene anomalies can be described as anomalies in crowded scenes with a large number of humans \cite{Zeng2023}. It is comparatively difficult to identify anomalous behaviours in dense scenes than sparse scenes due to tracking multiple people and finding their individual anomaly scores \cite{morais2019learning}. Thus, there is a need to develop methods that can effectively identify both sparse and dense scene anomalies.
With the development of algorithms for handling different types of anomalies, there is a need for datasets composed of the specific type of anomalies to ensure efficient training and evaluation. This can be handled by either having separate datasets for specific types of anomalies or general datasets with a distribution of multiple types of anomalies.


\subsection{Hardware}
The skeletons collected using Microsoft Kinect (depth) camera has been used in the past studies \cite{nguyen2016skeleton, baptista2018deformation}. However, the defunct production of the Microsoft Kinect camera \cite{kinect} has led to hardware constraints in the further development of skeletal anomaly detection approaches. Other commercial products include Vicon \cite{vicon2019} with optical sensors and TheCaptury \cite{captury2019} with multiple cameras. But they function in very constrained environments or require special markers on the human body. New cameras, such as `Sentinare 2' from AltumView \cite{altumview}, circumvent such hardware requirements by directly processing videos on regular RGB cameras and transmitting skeletons information in real-time.

\subsection{Tracking skeletons}
The existing approaches for skeletal video anomaly detection involve spatio-temporal skeleton graphs \cite{luo2021normal} or temporal sequences \cite{morais2019learning}, which are constructed by tracking an individual across multiple frames. However, this is challenging in scenarios where there are multiple people within a scene. The entry and exit of people in the scene, overlapping of people during movement and presence of occluding objects make tracking people across frames a very challenging task. 

\subsection{Choice of threshold}
There can be deployment issues in these methods because the choice of threshold is not clear. In the absence of any validation set (containing both normal and unseen anomalies) in an anomaly detection setting, it is very hard to fine-tune an operating threshold using just the training data (comprising of normal activities only). To handle these situations, outliers within the normal activities can be used as a proxy for unseen anomalies \cite{khan2017detecting,khanempirical}; however, inappropriate choices can lead to increased false alarms or missed alarms. Domain expertise can be utilized to adjust a threshold, which may not be available in many cases.

\subsection{Decision granularity}
There is a need to address the challenges associated with the granularity and the decision-making time of the skeletal video anomaly detection methods for real-time applications. The existing methods mostly output decisions on a frame level, which becomes an issue when the input to the method is a real-time continuous video stream at multiple frames per second. This can lead to alarms going off multiple times a second, which can be counter-productive. One solution is for the methods to make decisions on a time-window basis, each window of length of a specified duration. However, this brings in the question about the optimal length of each decision window. A short window is impractical as it can lead to frequent and repetitive alarms, while a long window can lead to missed alarms, and delayed response and intervention. Domain knowledge can be used to make a decision about the length of decision windows.

\section{Future Directions}
Skeletons can be used in conjunction with optical flow \cite{duman2019anomaly} to develop privacy-protecting approaches to jointly learn from temporal and structural modalities. Approaches based on federated learning (that do not combine individual data, but only the models) can further improve the privacy of these methods \cite{abedi2020fedsl}. Segmentation masks \cite{yan2020image} can be leveraged in conjunction with skeletons to occlude humans while capturing the information pertaining to scene and human motion to develop privacy-protecting anomaly detection approaches. 

The skeletons signify motion and posture information for the individual humans in the video; however, they lack information regarding human-human and human-object interactions. Information pertaining to interaction of the people with each other and the objects in the environment is important for applications such as, violence detection \cite{boekhoudt2021hr}, theft detection \cite{boekhoudt2021hr} and agitation detection \cite{9684388} in care home settings. Skeletons can be used to replace the bodies of the participants, while keeping the background information in video frames \cite{mishra2022privacy} to analyze both human-human and human-object interaction anomalies.
Further, object bounding boxes can be used in conjunction with human skeletons to model human-object interaction while preserving the privacy of humans in the scene. The information from other modalities (e.g. wearable devices) along with skeleton features can be used to develop multi-modal anomaly detection methods to improve the detection performance. Further, the generated embeddings of relevant supervised approaches \cite{zheng20213d, zhang2022actionformer} can be used to fine tune skeletal video anomaly detection models.

As can be seen in Table \ref{tab:tab1} and \ref{tab:tab1_cotd}, the existing skeletal video anomaly detection methods and available datasets focus towards detecting irregular body postures \cite{luo2021normal}, and anomalous human actions \cite{pang2022predicting} in mostly outdoor settings, and not in proper healthcare settings, such as personal homes and long-term care homes. This a gap towards real world deployment, as there is a need to extend the scope of detecting anomalous behaviours using skeletons to in-home and care home settings, where privacy is a very important concern. This can be utilized to address important applications, such as fall detection \cite{feng2014fall}, agitation detection \cite{9684388, mishra2022privacy}, and independent assistive living. This will help to develop supportive homes and communities and encourage autonomy and independence among the increasing older population and dementia residents in care homes. While leveraging skeletons helps to get rid of facial identity and appearance-based information, it is important to ask the question if skeletons can be considered private enough \cite{wang2018learning, liao2020model} and what steps can be taken to further anonymize the skeletons. 
Another potential area of investigation for real-world deployment of privacy-protecting anomaly detection systems would be to perform video data acquisition, skeletal tracking (e.g., MediaPipe \cite{lugaresi2019mediapipe}) and model inferencing in real-time. However, there may be challenges around integrating cloud services, on-chip embedding of AI algorithms, the latency of reaction time, internet stability and false positive rates.

\section{Conclusion}
In this paper, we provided a survey of recent works that leverage the skeletons or body joints estimated from videos for the anomaly detection task. 
The skeletons hide the facial identity and overall appearance of people and can provide vital information about joint angles \cite{guoexercise}, speed of walking \cite{kovavc2014human}, and interaction with other people in the scene \cite{morais2019learning}. 
Our literature review showed that many deep learning-based approaches leverage reconstruction, prediction error and their other combinations to successfully detect anomalies in a privacy protecting manner. This review suggests the first steps towards increasing adoption of devices (and algorithms) focused on improving privacy in a residential or communal setting. It will further improve the deployment of anomaly detection systems to improve the safety and care of people. 
The skeleton-based anomaly detection methods can be used to design privacy-preserving technologies for the assisted living of older adults in a care environment \cite{chaaraoui2012review} 
or enable older adults to live independently in their own homes to cope with the increasing cost of long-term care demands \cite{hbali2018skeleton}. 
Privacy-preserving methods using skeleton features can be employed to assist with skeleton-based rehab exercise monitoring \cite{obdrvzalek2012accuracy} or in social robots for robot-human interaction \cite{garcia2019human} that assist older people in their activities of daily living.

\section{Acknowledgements}
This work was supported by AGE-WELL NCE Inc, Alzheimer's Association, Natural Sciences and Engineering Research Council and UAE Strategic Research Grant.

\bibliographystyle{IEEEtran}
\bibliography{ms}

\vspace{-25pt}
\begin{IEEEbiography}[{\includegraphics[width=1in,height=1.25in,clip,keepaspectratio]{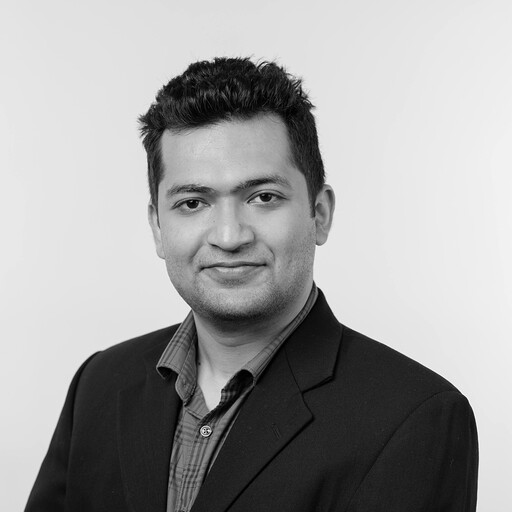}}]{Pratik K. Mishra} obtained his Masters in Computer Science and Engineering from the Indian Institute of Technology (IIT) Indore, India, in 2020. He is currently pursuing his Ph.D. from the Institute of Biomedical Engineering, University of Toronto and working towards the application of computer vision for detecting behaviours of risk in people with dementia. Previously, he worked as a research volunteer at the Toronto Rehabilitation Institute, Canada and as a Data Management Support Specialist at IBM India Private Limited. 
\end{IEEEbiography}

\vspace{-36pt}
\begin{IEEEbiography}[{\includegraphics[width=1in,height=1.25in,clip,keepaspectratio]{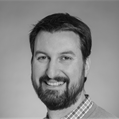}}]{Alex Mihailidis}, PhD, PEng, is the Barbara G. Stymiest Research Chair in Rehabilitation Technology at KITE Research Institute at University Health Network/University of Toronto. He is the Scientific Director of the AGE-WELL Network of Centres of Excellence, which focuses on the development of new technologies and services for older adults. He is a Professor in the Department of Occupational Science and Occupational Therapy and in the Institute of Biomedical Engineering at the University of Toronto (U of T), as well as holds a cross appointment in the Department of Computer Science at the U of T. 

Mihailidis is very active in the rehabilitation engineering profession and is the Immediate Past President for the Rehabilitation Engineering and Assistive Technology Society for North America (RESNA) and was named a Fellow of RESNA in 2014, which is one of the highest honours within this field of research and practice. His research disciplines include biomedical and biochemical engineering, computer science, geriatrics and occupational therapy. Alex is an internationally recognized researcher in the field of technology and aging. He has published over 150 journal and conference papers in this field and co-edited two books: Pervasive computing in healthcare and Technology and Aging.
\end{IEEEbiography}

\vspace{-36pt}
\begin{IEEEbiography}[{\includegraphics[width=1in,height=1.25in,clip,keepaspectratio]{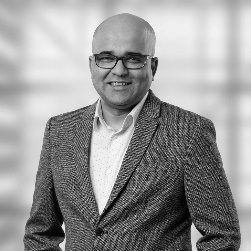}}]{Shehroz S. Khan} obtained his B.Sc Engineering, Masters and Phd degrees in computer science in 1997, 2010 and 2016. He is currently working as a Scientist at KITE – Toronto Rehabilitation Institute (TRI), University Health Network, Canada. He is also cross appointed as an Assistant Professor at the Institute of Biomedical Engineering, University of Toronto (UofT). Previously, he worked as a postdoctoral researcher at the UofT and TRI. Prior to joining academics, he worked in various scientific and researcher roles in the industry and government jobs. He is an associate editor of the Journal of Rehabilitation and Assistive Technologies. He has organized four editions of the peer-reviewed workshop on AI in Aging, Rehabilitation and Intelligent Assisted Living held with top AI conferences (ICDM and IJCAI) from 2017-2021. His research is funded through several granting agencies in Canada and abroad, including NSERC, CIHR, AGEWELL, SSHRC, CABHI, AMS Healthcare, JP Bickell Foundation, United Arab Emirates University and LG Electronics.  He has published $49$ peer-reviewed research papers and his research focus is the development of AI algorithms for solving aging related health problems. 
\end{IEEEbiography}

\end{document}